%% file: main.tex
\documentclass[10pt,twocolumn,letterpaper]{article}

 \usepackage[pagenumbers]{iccv} 


\definecolor{iccvblue}{rgb}{0.21,0.49,0.74}
\usepackage[pagebackref,breaklinks,colorlinks,allcolors=iccvblue]{hyperref}
\usepackage{multirow}
\usepackage{array}
\usepackage{colortbl}
\usepackage{booktabs}
\usepackage{bm}
\usepackage{float}
\usepackage{graphicx}
\usepackage{caption}

\def\paperID{13415} 
\def\confName{ICCV}
\def\confYear{2025}

\title{Language-Guided Visual Perception Disentanglement for Image Quality Assessment and Conditional Image Generation}

\author{Zhichao Yang\textsuperscript{1} \quad Leida Li\textsuperscript{1,2}\thanks{Corresponding author} \quad Pengfei Chen\textsuperscript{1} \quad Jinjian Wu\textsuperscript{1} \quad Giuseppe Valenzise\textsuperscript{3}\\
{\textsuperscript{1}Xidian University, \textsuperscript{2}Chongqing Three Gorges University, \textsuperscript{3}Paris-Saclay University}\\
{\tt\small yangzhichao@stu.xidian.edu.cn, (ldli,chenpengfei,jinjian.wu)@xidian.edu.cn,} \\
{\tt\small giuseppe.valenzise@l2s.centralesupelec.fr}
}

\begin{document}
\maketitle
\input{sec/0_abstract}
\input{sec/1_intro}
\input{sec/2_relatedwork}

\input{sec/3_method}

\input{sec/4_experiment}
\input{sec/5_conclusion}
{
    \small
    \bibliographystyle{ieeenat_fullname}
    \bibliography{ref}
}

\end{document}

%% file: sec/0_abstract.tex
\begin{abstract}
Contrastive vision-language models, such as CLIP, have demonstrated excellent zero-shot capability across semantic recognition tasks, mainly attributed to the training on a large-scale I\&1T (one Image with one Text) dataset. This kind of multimodal representations often blend semantic and perceptual elements, placing a particular emphasis on semantics. However, this could be problematic for popular tasks like image quality assessment (IQA) and conditional image generation (CIG), which typically need to have fine control on perceptual and semantic features. Motivated by the above facts, this paper presents a new multimodal disentangled representation learning framework, which leverages disentangled text to guide image disentanglement. To this end, we first build an \textbf{I\&2T} (one Image with a perceptual Text and a semantic Text) dataset, which consists of disentangled perceptual and semantic text descriptions for an image. Then, the disentangled text descriptions are utilized as supervisory signals to disentangle pure perceptual representations from CLIP's original `coarse' feature space, dubbed \textbf{DeCLIP}. Finally, the decoupled feature representations are used for both image quality assessment (technical quality and aesthetic quality) and conditional image generation. Extensive experiments and comparisons have demonstrated the advantages of the proposed method on the two popular tasks. The dataset, code, and model will be available.
\end{abstract}

%% file: sec/1_intro.tex
\section{Introduction}

\begin{figure}[t]
	\centering
	\includegraphics[width=\linewidth]{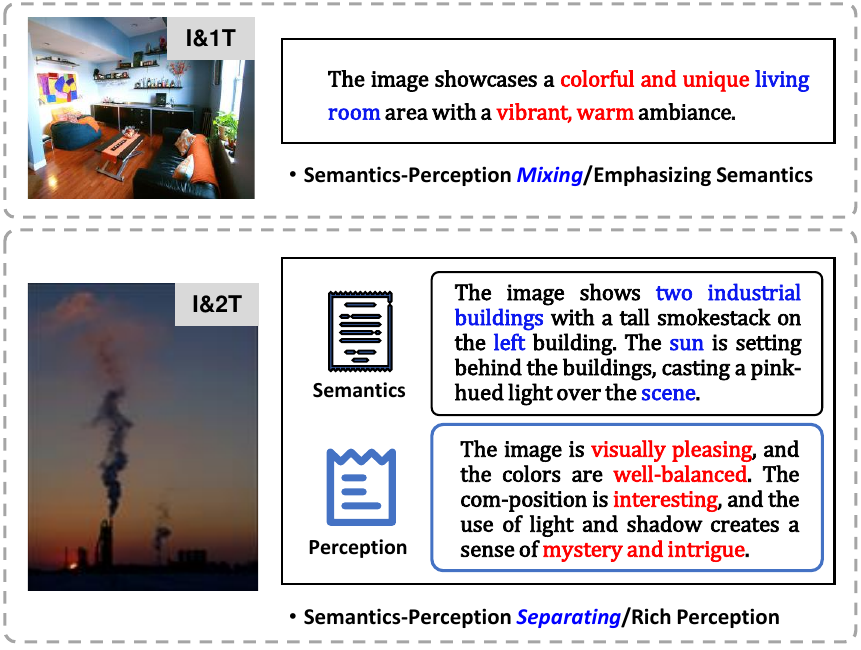}
	
	\caption{Conceptual difference between I\&1T and I\&2T. Upper: `{I\&1T}' features an image accompanied by one description that blends semantic and perceptual elements, with focus on semantics. Bottom: `{I\&2T}' features an image paired with two descriptions, offering rich perceptual insights while clearly differentiating between perception and semantics.}
	\label{fig1}
\end{figure}

Semantics and perception are two basic perspectives for understanding an image. Compared to the concretization and objectivity of semantics, perception exhibits inherent abstraction and subjectivity \cite{fang2020perceptual,lu2015deep}. The learning of high-quality perceptual representations from images has been a longstanding research topic in the computer vision community, with extensive applications including image recommendation \cite{ren2017personalized,yang2023multi}, image enhancement \cite{wang2019progressive,lu2019unsupervised}, and style transfer \cite{gatys2017controlling,lee2018diverse}, etc.

In the past few years, significant efforts have been dedicated to the representation and extraction of perceptual features, encompassing aspects such as visual quality \cite{zhu2020metaiqa,ke2021musiq,zhang2023blind} and aesthetic appeal \cite{talebi2018nima,li2020personality,ke2023vila}. Early methodologies primarily relied on hand-crafted features, necessitating a substantial amount of domain expertise \cite{datta2006studying,ke2006design}. In the deep learning era, Convolutional Neural Networks (CNNs) and Transformers have been commonly employed to uncover the perceptual representations embedded within large-scale datasets \cite{hosu2020koniq,murray2012ava}. However, such methods rely heavily on human perceptual annotations, typically in the form of Mean Opinion Score (MOS). Recent achievements in multi-modal methods have integrated the language modality to enhance the understanding of visual modality. The perceptual representation of the language modality has received considerable attention and investigation, significantly enhancing the perceptual understanding of the visual modality \cite{ke2023vila,sheng2023aesclip}.

Using image-text pairs with a cross-modal contrastive loss proves to be highly effective for multi-modal representation learning \cite{yuan2021multimodal}. Contrastive Language-Image Pre-Training, i.e. CLIP \cite{radford2021learning}, trained on a billion-level {I\&1T} (one Image with one Text) dataset, demonstrates remarkable zero-shot generalization capability across a wide range of downstream tasks. However, this kind of text description typically mixes the semantic and perceptual aspects of an image, with more emphasis on semantics, as illustrated in the upper part of \cref{fig1}, which limits model's perceptual representation capability. In this context, researchers have made significant effort in establishing datasets rich in perceptual descriptions \cite{wu2024q,huang2024aesexpert}. However, the entanglement between semantics and perception in the visual and language modalities remains challenging. This leads to significant limitations on the generalization capability of models in Image Quality Assessment (IQA) and multi-modal Conditional Image Generation (CIG). For IQA task, diversified semantic backgrounds poses a significant challenge to the evaluation capability of models, especially in cross-scene and cross-dataset contexts. For CIG task, the intricate coupling of semantics and perception within image prompts hinders generative models from obtaining accurate references, thereby affecting the final generation quality.

The perceptual description of an image, encompassing aspects such as clarity, light, composition etc., differs markedly from the corresponding semantic description, which pertains to scenes and objects, as illustrated in the bottom part of \cref{fig1}. Compared to the intricate entanglement between semantics and perception in the visual modality, the two aspects are relatively easy to differentiate in the language modality, which offers a natural form of decoupled supervision. Motivated by the above facts, this paper presents a novel multi-modal framework of Decoupled-Text Guided Image Disentangled Representation Learning for accurate vision-language alignment. First, we establish a dataset for multi-modal disentangled representation learning, called \textbf{{I\&2T}} (one Image with a perceptual Text and a semantic Text), which features images with distinct perceptual and semantic descriptions. Concretely, we utilize existing multi-modal data with rich perceptual descriptions, leveraging the strong semantic understanding and summarization capabilities of Multi-modal Large Language Models (MLLMs, e.g., GeminiPro \cite{Gemini}) to design prompts for data generation. Building upon this, we explore the use of decoupled descriptions to refine CLIP's original vision-language alignment space, dubbed \textbf{DeCLIP}. In particular, based on the original robust yet `coarse' vision-language space of CLIP, we leverage decoupled text descriptions as supervision to learn `pure' perceptual and semantic representations of the visual modality. DeCLIP exhibits enhanced zero-shot generalization capability in image quality assessment, including technical quality and aesthetic quality. Moreover, we propose a method that enables nuanced control over the perceptual and semantic aspects of image generation by employing Decoupled vision Prompts to Adapt the vanilla Stable Diffusion \cite{rombach2022high}, referred to as \textbf{DP-Adapter}.

We summarize the contributions of this work as follows:

\begin{itemize}
	\item {We present a new multi-modal framework of Decoupled-Text Guided Image Disentangled Representation Learning, referred to as DeCLIP, which can effectively disentangle the latent perceptual and semantic factors in pre-trained CLIP-like models, enabling more accurate and fine-grained vision-language alignment.}
	\item {We construct an {I\&2T} dataset for multi-modal disentangled representation learning, which includes images accompanied by distinct perceptual and semantic descriptions. These decoupled descriptions provide supervision for the disentanglement of the visual modality.}
	\item {We utilize the decoupled alignment of vision-language representations from DeCLIP for two popular tasks: Image Quality Assessment and Conditional Image Generation. Extensive experiments and comparisons demonstrate that the model achieves excellent zero-shot generalization capability in both tasks.}
\end{itemize}

%% file: sec/2_relatedwork.tex
\section{Related Work}

\begin{figure*}[ht]
	\centering
	\includegraphics[width=0.89\linewidth]{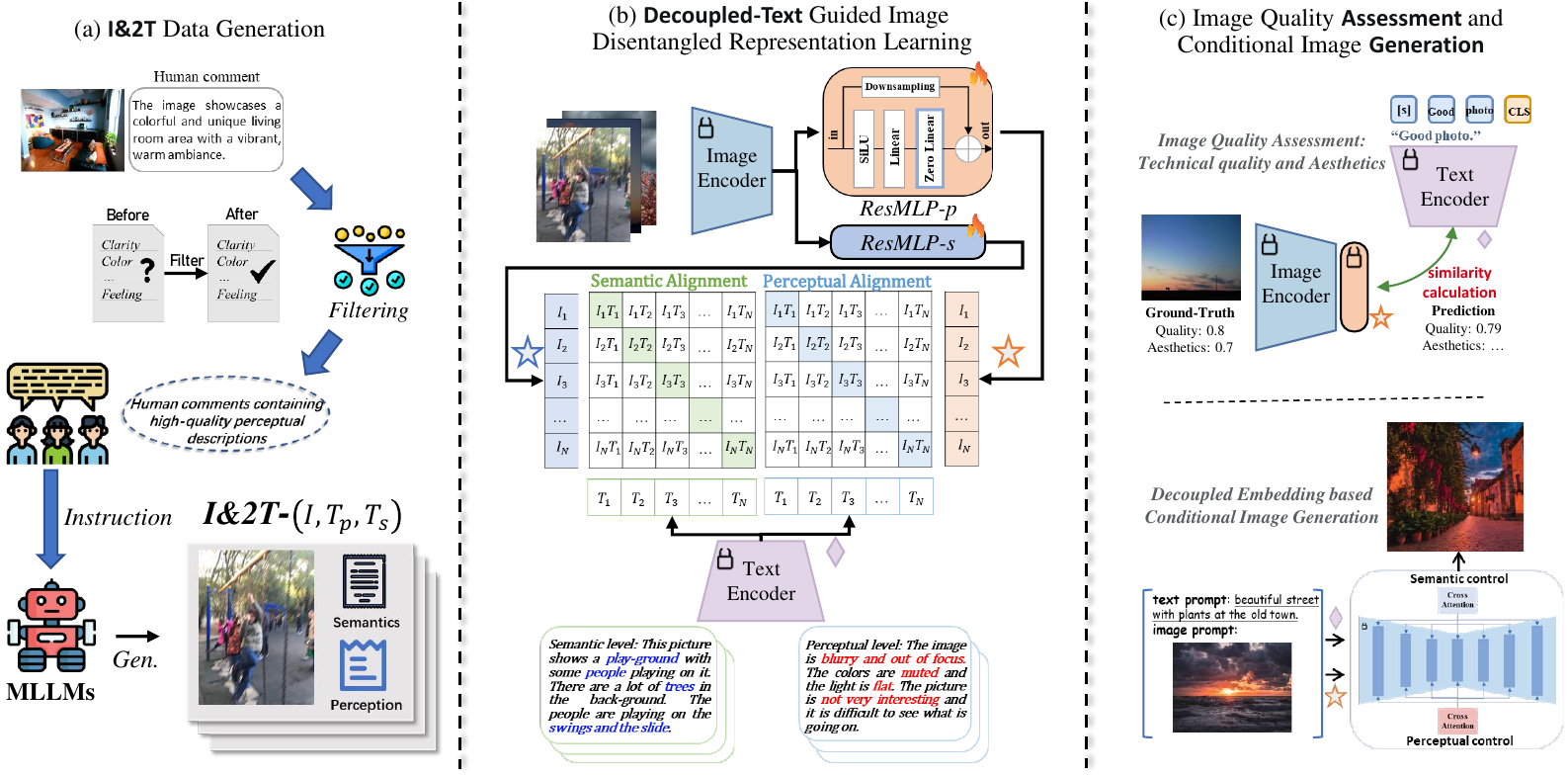}
	
	\caption{The overall framework of the proposed method. First, the {I\&2T} dataset, consisting of 112,769 image-text pairs formatted as \( \left ( I, T_{p}, T_{s} \right ) \), is established. Building upon this, decoupled text is utilized as supervision to disentangle pure semantic and perceptual features. Finally, the resulting decoupled representations are employed for image quality assessment and conditional image generation. }
	\label{fig2}
\end{figure*}

\subsection{Image Quality Assessment}
Image quality assessment aims to assess the subjective properties of images consistent with human perception \cite{hagtvedt2008perception}, including technical quality \cite{8576582,zhu2020metaiqa,su2020blindly,ke2021musiq,wang2023exploring,zhang2023blind} and aesthetic quality \cite{talebi2018nima,li2020personality,he2022rethinking,10054147,ke2023vila,yang2024semantics}. Early methods designed hand-crafted features to represent the perceptual characteristics of images, necessitating a substantial amount of domain expertise \cite{tang2013content}. Recently, deep learning-based works focused on data-driven methods, which are typically based on large-scale datasets containing images and human ratings \cite{hosu2020koniq,talebi2018nima}. More recently, the advancements in multi-modal vision have prompted researchers to focus on the perceptual description in the language modality \cite{zhang2023blind,yang2024semantics}. However, the intricate entanglement of semantics and perception in the visual modality remains a significant obstacle in enhancing the perceptual capabilities, leading to the inherent weakness in generalization of quality evaluation models. In this paper, we explore the use of decoupled text descriptions to disentangle pure perceptual representations from the visual modality, thereby enhancing the zero-shot generalization capability in image quality assessment.
\subsection{Multi-modal Conditional Image Generation}
Recent advancements in multi-modal conditional image generation often leverage pre-trained vision-language models like CLIP \cite{radford2021learning}, alongside large diffusion models like Stable Diffusion \cite{rombach2022high}. CLIP bridges the gap between vision and language, while Stable Diffusion functions as a latent diffusion model trained on a vast dataset. IP-Adapter \cite{ye2023ip} utilized a decoupled cross-attention mechanism to connect CLIP with pretrained text-to-image diffusion models, enabling image prompt capability. However, achieving accurate control over the semantic and perceptual aspects of generated images is extremely challenging with CLIP's coarse alignment space. To address this issue, this paper presents the decoupled vision-language representations as prompts to enhance conditional image generation.
\subsection{Disentangled Representation Learning}
Disentangled representation learning aims to isolate intrinsic latent factors within data, transforming them into distinct and controllable representations \cite{wang2024disentangled}. In the context of single modality, some studies utilized image augmentation techniques to separate content variables from the latent space by inducing significant perceptual changes \cite{zhao2023quality,venkataramanan2024joint}. For multi-modal scenarios, CLAP \cite{caiclap} employed text augmentation with a cross-modal contrastive loss to disentangle latent content variables. In contrast to these approaches, the proposed method utilizes decoupled semantic and perceptual descriptions to disentangle visual modality.

%% file: sec/3_method.tex
\section{Proposed Method}
In this section, we provide a detailed introduction to the proposed approach. The pipeline is illustrated in \cref{fig2}. First, the {I\&2T} dataset, consisting of images paired with distinct perceptual and semantic descriptions, is established. Building upon this, the decoupled text is utilized as supervision to disentangle pure perceptual and semantic features from the visual modality. Finally, the disentangled visual representations are employed for image quality assessment and conditional image generation.

\begin{table}[htbp]
	\centering
	\small
	\setlength{\tabcolsep}{4pt}
	\begin{tabular}{ccc}
		\toprule
		Data Source & Filtered Quantity & Image Types \\ 
		\midrule
		AesExpert \cite{huang2024aesexpert}                  & 21390                 & Natural, Art, AIGC     \\
		Q-Instruct \cite{wu2024q}               & 11753                 & Natural, AIGC          \\         
		ShareGPT4v \cite{chen2023sharegpt4v}               & 22643                 & Natural, Art           \\
		AVA-Comments \cite{zhou2016joint}            & 56983                 & Natural                \\ 
		\midrule
		\rowcolor[cmyk]{.1,0,0,0} 
		{I\&2T}                      & 112,769               & Natural, Art, AIGC     \\ 
		\bottomrule
	\end{tabular}
	\caption{Overview of the data source for {I\&2T}. The filtered quantity and image types are also presented.}
	\label{tab1}
\end{table}

\begin{figure}[t]
	\centering
	\includegraphics[width=0.97\linewidth]{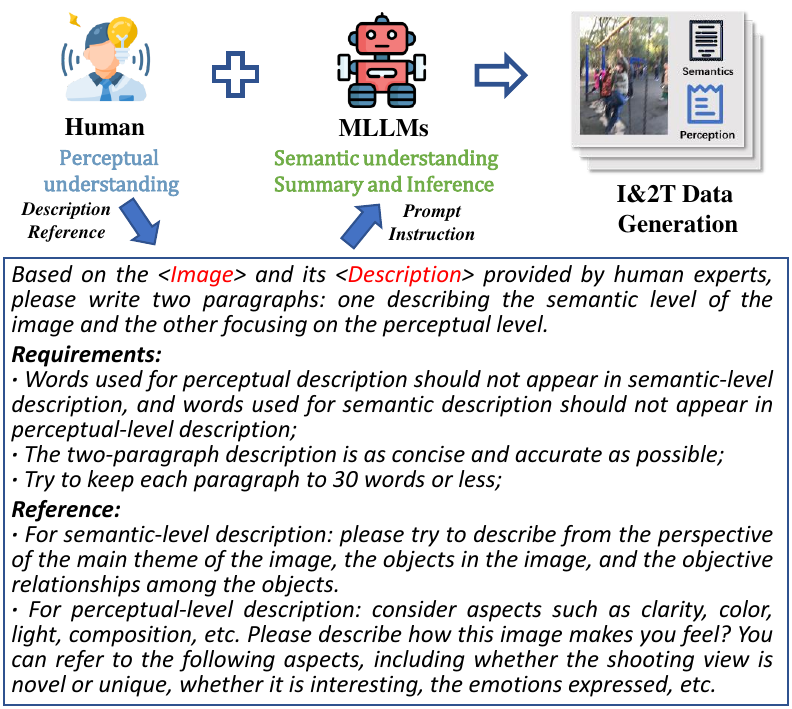}	
	\caption{The I\&2T database construction pipeline and the prompt design. The perceptual abilities of humans, in conjunction with the semantic understanding and reasoning capabilities of MLLMs, collaboratively contribute to data generation. }
	\label{fig5}
\end{figure}

\subsection{I\&2T Data Generation}
Compared to the intricate entanglement of semantics and perception in the visual modality, the language modality is inherently easier to disentangle \cite{cheng2024disentangled}. Words that describe semantics are frequently linked to objects or scenes, whereas words that pertain to perception are often associated with attributes, such as clarity, composition, and emotional resonance. Recently, Zhong \emph{et al}. \cite{zhong2023aesthetically} conducted a manual selection of object-related and aesthetics-related words to compute the Aesthetic Relevance Score (ARS) for sentences. Yang \emph{et al}. \cite{yang2024semantics} employed semantic-related and attribute-related tags to extract corresponding features for evaluating image aesthetics. Inspired by these works, we construct the {I\&2T} dataset, which features images with distinct perceptual and semantic descriptions.

First, we select representative perception-related multi-modal dataset (e.g., AVA-Comments \cite{zhou2016joint}), as well as datasets used for training perceptual MLLMs (e.g., Q-Instruct \cite{wu2024q}, AesExpert \cite{huang2024aesexpert}, ShareGPT4v \cite{chen2023sharegpt4v}), to serve as data sources. These datasets contain rich human-generated perceptual descriptions, but the quality varies significantly. Therefore, we employ the library rich in perceptual vocabulary \cite{yang2024semantics} to filter these datasets. The number of perceptual vocabularies included in each description serves as selection criterion. We exclude data containing fewer than seven perceptual vocabulary items, which is considered as low-quality perceptual description. Finally, we collect 112,769 image-text pairs as the initial data, including natural, art and AIGC types, as illustrated in \cref{tab1}.

Compared to perceptual understanding, the semantic understanding of general Multi-modal Large Language Models (MLLMs) is more reliable \cite{wu2023q,huang2024aesbench}. Consequently, we employ high-quality human-generated perceptual descriptions as instructions to craft prompts that facilitate the MLLMs in re-labeling decoupled semantic and perceptual descriptions for each image. Concretely, we formulate specific descriptive requirements and references to guide MLLMs in generating decoupled descriptions. The complete prompt and the construction pipeline are illustrated in \cref{fig5}. 

The final {I\&2T} dataset can be represented as follows: 
\begin{equation}
	\mathcal D_{I\&2T}=\left \{ \left ( I^{i}, T_{p}^{i}, T_{s}^{i}   \right )  \right \} _{i=1}^{N },
	\label{eq:1}
\end{equation}
where $I^{i}$ denotes the \(i\)-th image, $\left ( T_{p}^{i}, T_{s}^{i}  \right ) $ denotes the corresponding perceptual and semantic descriptions, and $N$ represents the number of images.

\subsection{Decoupled-Text Guided Image Disentangled Representation Learning}
Building upon the {I\&2T} dataset, we utilize decoupled textual descriptions as supervision to refine vision-language alignment of CLIP, thereby achieving disentangled perceptual and semantic representations in the visual modality, dubbed as DeCLIP.

\textbf{Preliminary of CLIP:}
CLIP is a Contrastive Language-Image Pretraining model that establishes a joint embedding space for images and texts. It consists of an image encoder and a text encoder, trained using a contrastive learning framework on a large-scale I\&1T dataset.

\textbf{Encoding Image:}
Compared to the strong semantic alignment capabilities, CLIP's perceptual abilities are relatively weaker. However, previous studies \cite{hentschel2022clip} and \cite{wang2023exploring} have demonstrated the potential of CLIP in evaluating both aesthetics and technical quality of images. Furthermore, many low-level visual tasks, such as image restoration \cite{xu2024towards} and image enhancement \cite{liang2023iterative}, have utilized CLIP's perceptual understanding. These investigations highlight the substantial perceptual understanding potential within CLIP's latent alignment space. Consequently, we introduce two simple but effective projectors on top of the existing image encoder in CLIP for disentanglement: $ResMLP_{p}$ and $ResMLP_{s}$ \cite{caiclap}, which consist of a residual block featuring a zero-initialized, bias-free linear layer. The zero-linear components learn specific knowledge for perceptual alignment and semantic alignment respectively, while the residual connection enables learning directly from the CLIP's pre-trained representation space, avoiding a random starting point. Specifically, the \(i\)-th image from the {I\&2T} dataset, is mapped to two embedding spaces, as follows:
\begin{equation}
	\boldsymbol f_{I}^{i}  = E_{I}\left ( I^{i}  \right ),
	\label{eq:2}
\end{equation}
\begin{equation}	
		\boldsymbol f_{I\left ( p \right )}^{i} = ResMLP_{p} \left ( \boldsymbol f_{I}^{i}  \right ), \boldsymbol f_{I\left ( s \right )}^{i} = ResMLP_{s} \left ( \boldsymbol f_{I}^{i}  \right ) ,
	\label{eq:3}
\end{equation}
where $\boldsymbol f_{I\left ( p \right )}^{i}, \boldsymbol f_{I\left ( s \right )}^{i}$ represent mapped visual embeddings corresponding to perception and semantics, respectively.

\textbf{Encoding Text:}
Considering the differences in terminology and focus between semantic and perceptual descriptions, we directly employ CLIP's text encoder $E_{T}$ to obtain distinct textual representations. Concretely, the perceptual and semantic texts corresponding to the \(i\)-th image, are inputted to generate decoupled embeddings:
\begin{equation}
	\begin{split}
		\boldsymbol f_{T\left ( p \right )}^{i} = E_{T} \left ( T_{p}^{i}  \right ),\boldsymbol f_{T\left ( s \right )}^{i} = E_{T} \left ( T_{s}^{i}  \right ),
	\end{split}
	\label{eq:4}
\end{equation}
where $\boldsymbol f_{T\left ( p \right )}^{i},\boldsymbol f_{T\left ( s \right )}^{i}$ denote textual embeddings associated with perception and semantics, respectively.

\textbf{Disentangled Representation Learning:}
Based on CLIP's original joint embedding space for images and texts, we leverage the decoupled texts of the {I\&2T} dataset to achieve decoupled vision-language alignment. Specifically, we align the decoupled perceptual and semantic textual embeddings with their corresponding visual representations. This transforms the original unified alignment space into two distinct spaces: one for perceptual alignment and the other for semantic alignment. This process is achieved through two cross-modal contrastive losses:
\begin{equation}
	\mathcal L^{Con}= \mathcal L_{p}^{Con}+\mathcal L_{s}^{Con},
	\label{eq:5}
\end{equation}
\begin{equation}
	\mathcal L_{p}^{Con}=-\frac{1}{B}\sum_{i}^{B}\log_{}{\frac{\exp\left ( \boldsymbol f_{I\left ( p \right ) }^{i} \cdot \boldsymbol f_{T\left ( p \right ) }^{i} / \tau  \right ) }{\sum_{j=1}^{B} \exp\left ( \boldsymbol f_{I\left ( p \right ) }^{i} \cdot \boldsymbol f_{T\left ( p \right ) }^{j} / \tau \right ) } }   ,
	\label{eq:6}
\end{equation}
\begin{equation}
	\mathcal L_{s}^{Con}=-\frac{1}{B}\sum_{i}^{B}\log_{}{\frac{\exp\left ( \boldsymbol f_{I\left ( s \right ) }^{i} \cdot \boldsymbol f_{T\left ( s \right ) }^{i} / \tau  \right ) }{\sum_{j=1}^{B} \exp\left ( \boldsymbol f_{I\left ( s \right ) }^{i} \cdot \boldsymbol f_{T\left ( s \right ) }^{j} / \tau \right ) } }   ,
	\label{eq:7}
\end{equation}
where $\mathcal L_{p}^{Con}$ is the image-to-perception contrastive loss, $\mathcal L_{s}^{Con}$ is the image-to-semantics contrastive loss, $\tau$ is the temperature coefficient, $\left ( \cdot  \right )  $ denotes the inner product operation, and $B$ is the batch size. It is essential to note that optimization is confined solely to the parameters of the two $ResMLP$ modules.

\begin{table*}[t]
	\centering
		\small
	\setlength{\tabcolsep}{1pt}
	\begin{tabular}{c||ccccccccccccccccccccccc}
		\toprule
		
		\multirow{7}{*}{\normalsize Quality}& \multicolumn{1}{c|}{Dist Type} &
		\multicolumn{11}{c|}{Synthetic} &
		\multicolumn{8}{c|}{Authentic} &
		\multicolumn{3}{c}{AIGC} \\  \cmidrule{2-24}
		& \multicolumn{1}{c|}{Databases} &
		\multicolumn{3}{c|}{\textbf{LIVE}} &
		\multicolumn{2}{c|}{\textbf{CSIQ}} &
		\multicolumn{3}{c|}{\textbf{TID2013}} &
		\multicolumn{3}{c|}{\textbf{KADID}} &
		\multicolumn{2}{c|}{\textbf{LIVEC}} &
		\multicolumn{3}{c|}{\textbf{KonIQ}} &
		\multicolumn{3}{c|}{\textbf{SPAQ}} &
		\multicolumn{3}{c}{\textbf{AGIQA-3K}} \\ \cmidrule{2-24}
		& \multicolumn{1}{c|}{Metrics} &
		\small SRCC &
		\multicolumn{2}{c}{\small PLCC } &
		\small SRCC &
		\small PLCC &
		\multicolumn{2}{c}{\small SRCC} &
		\small PLCC &
		\small SRCC &
		\multicolumn{2}{c|}{\small PLCC} &
		\small SRCC &
		\small PLCC &
		\multicolumn{2}{c}{\small SRCC} &
		\small PLCC &
		\multicolumn{2}{c}{\small SRCC} &
		\multicolumn{1}{c|}{\small PLCC} &
		\multicolumn{2}{c}{\small SRCC} &
		\small PLCC \\ \cmidrule{2-24}
		& \multicolumn{1}{c|}{CLIP} &
		0.510 &
		\multicolumn{2}{c}{0.537} &
		0.504 &
		0.558 &
		\multicolumn{2}{c}{0.450} &
		0.497 &
		0.387 &
		\multicolumn{2}{c|}{0.392} &
		0.355 &
		0.318 &
		\multicolumn{2}{c}{0.327} &
		0.352 &
		\multicolumn{2}{c}{0.439} &
		\multicolumn{1}{c|}{0.381} &
		\multicolumn{2}{c}{0.308} &
		0.359 \\
		& \multicolumn{1}{c|}{{\begin{tabular}[c]{@{}c@{}}DeCLIP\\ ($\uparrow$\underline{\%})\end{tabular}}} &
		{\begin{tabular}[c]{@{}c@{}}0.725\\ {$\uparrow$\small\underline{21.5}}\end{tabular}} &
		\multicolumn{2}{c}{{\begin{tabular}[c]{@{}c@{}}0.707\\ {$\uparrow$\small\underline{17.0}}\end{tabular}}} &
		{\begin{tabular}[c]{@{}c@{}}0.569\\ {$\uparrow$\small\underline{6.5}}\end{tabular}} &
		{\begin{tabular}[c]{@{}c@{}}0.621\\ {$\uparrow$\small\underline{6.3}}\end{tabular}} &
		\multicolumn{2}{c}{{\begin{tabular}[c]{@{}c@{}}0.551\\ {$\uparrow$\small\underline{10.1}}\end{tabular}}} &
		{\begin{tabular}[c]{@{}c@{}}0.594\\ {$\uparrow$\small\underline{9.7}}\end{tabular}} &
		{\begin{tabular}[c]{@{}c@{}}0.502\\ {$\uparrow$\small\underline{11.5}}\end{tabular}} &
		\multicolumn{2}{c|}{{\begin{tabular}[c]{@{}c@{}}0.527\\ {$\uparrow$\small\underline{13.5}}\end{tabular}}} &
		{\begin{tabular}[c]{@{}c@{}}0.624\\ {$\uparrow$\small\underline{26.9}}\end{tabular}} &
		{\begin{tabular}[c]{@{}c@{}}0.513\\ {$\uparrow$\small\underline{19.5}}\end{tabular}} &
		\multicolumn{2}{c}{{\begin{tabular}[c]{@{}c@{}}0.602\\ {$\uparrow$\small\underline{27.5}}\end{tabular}}} &
		{\begin{tabular}[c]{@{}c@{}}0.593\\ {$\uparrow$\small\underline{24.1}}\end{tabular}} &
		\multicolumn{2}{c}{{\begin{tabular}[c]{@{}c@{}}0.793\\ {$\uparrow$\small\underline{35.4}}\end{tabular}}} &
		\multicolumn{1}{c|}{{\begin{tabular}[c]{@{}c@{}}0.446\\ {$\uparrow$\small\underline{6.5}}\end{tabular}}} &
		\multicolumn{2}{c}{{\begin{tabular}[c]{@{}c@{}}0.387\\ {$\uparrow$\small\underline{7.9}}\end{tabular}}} &
		{\begin{tabular}[c]{@{}c@{}}0.447\\ {$\uparrow$\small\underline{8.8}}\end{tabular}} \\ \midrule \midrule
		\multirow{7}{*}{\normalsize Aesthetics} & \multicolumn{1}{c|}{Img Type} &
		\multicolumn{11}{c|}{Natural} &
		\multicolumn{5}{c|}{Artistic} &
		\multicolumn{6}{c}{AIGC} \\ \cmidrule{2-24}
		& \multicolumn{1}{c|}{Databases} &
		\multicolumn{3}{c|}{\textbf{AVA}} &
		\multicolumn{2}{c|}{\textbf{AADB}} &
		\multicolumn{3}{c|}{\textbf{PARA}} &
		\multicolumn{3}{c|}{\textbf{EVA}} &
		\multicolumn{2}{c|}{\textbf{BAID}} &
		\multicolumn{3}{c|}{\textbf{APDD}} &
		\multicolumn{3}{c|}{\textbf{RichHF-18K}} &
		\multicolumn{3}{c}{\textbf{SAC}} \\ \cmidrule{2-24}
		& \multicolumn{1}{c|}{Metrics} &
		\small SRCC &
		\multicolumn{2}{c}{\small PLCC} &
		\small SRCC &
		\small PLCC &
		\multicolumn{2}{c}{\small SRCC} &
		\small PLCC &
		\small SRCC &
		\multicolumn{2}{c|}{\small PLCC} &
		\small SRCC &
		\small PLCC &
		\multicolumn{2}{c}{\small SRCC} &
		\multicolumn{1}{c|}{\small PLCC} &
		\multicolumn{2}{c}{\small SRCC} &
		\small PLCC &
		\multicolumn{2}{c}{\small SRCC} &
		\small PLCC \\ \cmidrule{2-24}
		& \multicolumn{1}{c|}{CLIP} &
		0.387 &
		\multicolumn{2}{c}{0.403} &
		0.334 &
		0.336 &
		\multicolumn{2}{c}{0.548} &
		0.550 &
		0.465 &
		\multicolumn{2}{c|}{0.464} &
		0.109 &
		0.084 &
		\multicolumn{2}{c}{0.319} &
		\multicolumn{1}{c|}{0.320} &
		\multicolumn{2}{c}{-0.078} &
		-0.057 &
		\multicolumn{2}{c}{0.241} &
		0.252 \\
		& \multicolumn{1}{c|}{{\begin{tabular}[c]{@{}c@{}}DeCLIP\\ ($\uparrow$\underline{\%})\end{tabular}}} &
		{\begin{tabular}[c]{@{}c@{}}0.395\\ {$\uparrow$\small\underline{0.8}}\end{tabular}} &
		\multicolumn{2}{c}{{\begin{tabular}[c]{@{}c@{}}0.409\\ {$\uparrow$\small\underline{0.6}}\end{tabular}}} &
		{\begin{tabular}[c]{@{}c@{}}0.403\\ {$\uparrow$\small\underline{6.9}}\end{tabular}} &
		{\begin{tabular}[c]{@{}c@{}}0.401\\ {$\uparrow$\small\underline{6.5}}\end{tabular}} &
		\multicolumn{2}{c}{{\begin{tabular}[c]{@{}c@{}}0.627\\ {$\uparrow$\small\underline{7.9}}\end{tabular}}} &
		{\begin{tabular}[c]{@{}c@{}}0.605\\ {$\uparrow$\small\underline{5.5}}\end{tabular}} &
		{\begin{tabular}[c]{@{}c@{}}0.506\\ {$\uparrow$\small\underline{4.1}}\end{tabular}} &
		\multicolumn{2}{c|}{{\begin{tabular}[c]{@{}c@{}}0.516\\ {$\uparrow$\small\underline{5.2}}\end{tabular}}} &
		{\begin{tabular}[c]{@{}c@{}}0.171\\ {$\uparrow$\small\underline{6.2}}\end{tabular}} &
		{\begin{tabular}[c]{@{}c@{}}0.148\\ {$\uparrow$\small\underline{6.4}}\end{tabular}} &
		\multicolumn{2}{c}{{\begin{tabular}[c]{@{}c@{}}0.411\\ {$\uparrow$\small\underline{9.2}}\end{tabular}}} &
		\multicolumn{1}{c|}{{\begin{tabular}[c]{@{}c@{}}0.388\\ {$\uparrow$\small\underline{6.8}}\end{tabular}}} &
		\multicolumn{2}{c}{{\begin{tabular}[c]{@{}c@{}}-0.018\\ {$\uparrow$\small\underline{6.0}}\end{tabular}}} &
		{\begin{tabular}[c]{@{}c@{}}-0.036\\ {$\uparrow$\small\underline{2.1}}\end{tabular}} &
		\multicolumn{2}{c}{{\begin{tabular}[c]{@{}c@{}}0.254\\ {$\uparrow$\small\underline{1.3}}\end{tabular}}} &
		{\begin{tabular}[c]{@{}c@{}}0.264\\ {$\uparrow$\small\underline{1.2}}\end{tabular}} \\ 
		\bottomrule
	\end{tabular}
	\caption{Comparison between the proposed DeCLIP and the vanilla CLIP (Zero-shot) across two perceptual quality evaluation tasks: Technical Quality and Aesthetics. The relative improvements are displayed.}
	\label{tab2}
\end{table*}

\subsection{Image Quality Assessment and Conditional Image Generation}
In this section, we utilize the decoupled vision-language features of DeCLIP for both Image Quality Assessment and Conditional Image Generation tasks.

\textbf{Image Quality Assessment:}
Compared to {I\&1T}, the {I\&2T} dataset offers rich perceptual descriptions that encompass various aspects, including visual quality and aesthetic appeal. To validate the effectiveness of the decoupled preceptual representations of images, we adopt a straightforward approach by evaluating DeCLIP's zero-shot capability across two popular perceptual quality evaluation tasks: Image Technical Quality Assessment (TQA) and Image Aesthetics Quality Assessment (AQA).

For TQA and AQA, we utilize the decoupled perceptual visual features in conjunction with antonym prompts (e.g., `Good photo.' and `Bad photo.') to predict image quality/aesthetics scores \cite{wang2023exploring}. Softmax function and cosine similarity are employed to implement this process:
\begin{equation}
	\bar{s} = \frac{e^{\cos \left ( \boldsymbol f_{I\left ( p \right ) }, \boldsymbol f_{TP}   \right ) } }{e^{\cos \left ( \boldsymbol f_{I\left ( p \right ) }, \boldsymbol f_{TP}   \right ) } + e^{\cos \left ( \boldsymbol f_{I\left ( p \right ) }, \boldsymbol f_{TN}   \right ) }} ,
	\label{eq:8}
\end{equation}
where $\bar{s}$ denotes the predicted score, $\cos \left ( \cdot  \right ) $ is the consine similarity function, and $\boldsymbol f_{TP}, \boldsymbol f_{TN}$ represent the positive and negative textual features of antonym prompts.

\begin{figure}[t]
	\centering
	\includegraphics[width=0.94\linewidth]{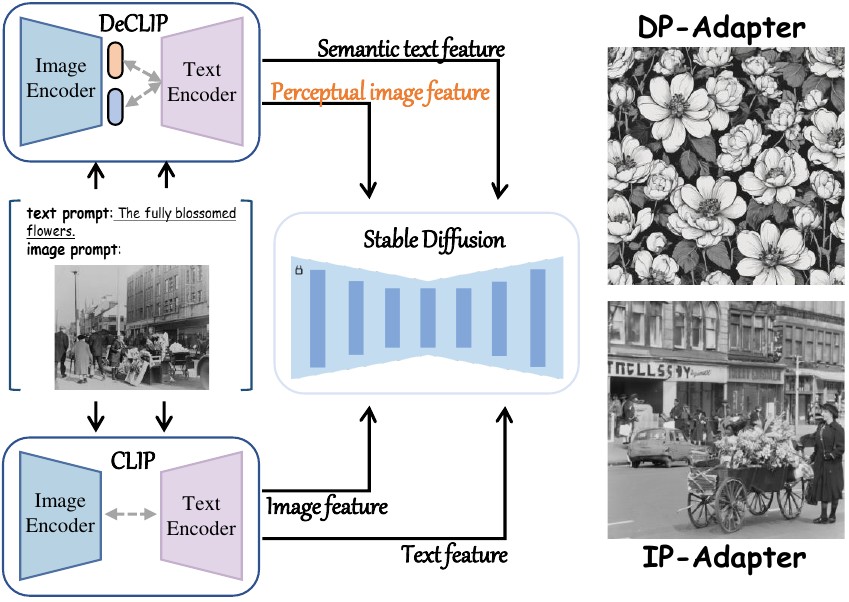}	
	\caption{DP-Adapter vs. IP-Adapter. The former utilizes the decoupled vision-language representations obtained from DeCLIP to control Stable Diffusion, while the latter directly employs CLIP's vision-language representations for generative control.}
	\label{fig3}
\end{figure}

\textbf{Conditional Image Generation:}
The image-text alignment capabilities of CLIP have been commonly applied in the realm of conditional image generation. IP-Adapter \cite{ye2023ip} presents a decoupled cross-attention mechanism based on CLIP for multi-modal conditional image generation. The training process can be expressed as:
\begin{equation}
	\mathcal L=\mathbb{E}_{x_{0}, \epsilon , c,t  }\left \| \epsilon -\epsilon _{\theta \left ( \boldsymbol x_{t} , c,t \right ) }  \right \|^{2}     ,
	\label{eq:10}
\end{equation}
where, a denoising process generates samples from Gaussian noise $\boldsymbol x_{t}\sim N\left ( 0,1 \right ) $ with a trained denoising model $\epsilon _{\theta \left ( x_{t} , c,t \right ) }$ parameterized by $\theta$. The $c$ represents the condition, which is generated by the text embeddings $\boldsymbol f_{T}$ and image embeddings $\boldsymbol f_{I}$ from CLIP, and integrated into the Stable Diffusion through the decoupled cross-attention module.

CLIP empowers the Stable Diffusion with the capability for multi-modal conditional image generation. Nevertheless, significant conflicts in prompts, such as overt semantic discrepancies, can adversely impact the quality of generated outputs, as illustrated at the bottom of \cref{fig3}. The decoupling of vision-language alignment spaces provides a viable solution to this challenge. By partitioning the original alignment space of CLIP into distinct perceptual and semantic sub-spaces, we can achieve flexible and more accurate control over the respective aspects of the generated images.

In particular, we utilize the decoupled vision-language representations of DeCLIP to adapt the Stable Diffusion for image generation. This enables flexible control over the perceptual and semantic aspects of generated images through the complementary use of image and text:
 \begin{equation}
 	c\in \left \{ \left ( \boldsymbol f_{I(p)}, \boldsymbol f_{T(s)}  \right ) , \left ( \boldsymbol f_{I(s)}, \boldsymbol f_{T(p)} \right )  \right \}      ,
 	\label{eq:11}
 \end{equation}
where $\left ( \boldsymbol f_{I(p)}, \boldsymbol f_{T(s)}  \right )$ and $\left ( \boldsymbol f_{I(s)}, \boldsymbol f_{T(p)} \right )$ represent two distinct control types: `perceptual image + semantic text' and `semantic image + perceptual text', respectively. An example is illustrated in the upper part of \cref{fig3}. These two modes better meet users' practical application needs, such as a user integrating perceptual elements from an aesthetically pleasing image into their desired semantic content for generation. It is worth noting that we do not retrain the Stable Diffusion and Attention module, because DeCLIP and CLIP essentially belong to the same alignment space.

%% file: sec/4_experiment.tex
\section{Experiment}
\subsection{Implementation Details}
We utilize ViT version of CLIP as the foundational model for disentangled representation learning. It is important to note that we only introduce two projectors during training, while maintaining all original settings of CLIP, including a fixed input size (e.g., 224×224) and the original parameters.  The projector comprises two linear layers and a SiLU activation function, with a latent dimension set to 1024. Data augmentation during training involves random cropping. For disentangled learning on the I\&2T dataset, we train with a batch size of 1024 for 200 epochs, employing a learning rate of 1e-5 along with a weight decay of 5e-2. The temperature coefficient $ \tau$ is set to 0.07. All experiments are conducted using the AdamW \cite{loshchilov2017decoupled} optimizer with four NVIDIA GeForce RTX 3090 GPUs.

\subsection{Performance on Image Quality Assessment}
\noindent{\textbf{Datasets and Metrics:}}
To evaluate the efficacy of the decoupled perceptual representations, we conduct experiments on two representative perceptual quality evaluation tasks: image technical quality assessment and image aesthetic quality assessment. For TQA, we test datasets with three different types of distortions, including synthetic distortions (LIVE \cite{sheikh2006statistical}, CSIQ \cite{larson2010most}, TID2013 \cite{ponomarenko2015image}, KADID \cite{lin2019kadid}), real-world distortions (LIVEC \cite{ghadiyaram2015massive}, KonIQ \cite{hosu2020koniq}, SPAQ \cite{fang2020perceptual}), and AI-generated distortions (AGIQA-3K \cite{li2023agiqa}). In terms of AQA, we test datasets with three distinct types of images: natural images (AVA \cite{murray2012ava}, AADB \cite{kong2016photo}, PARA \cite{yang2022personalized}, EVA \cite{kang2020eva}), artistic images (BAID \cite{yi2023towards}, APDD \cite{jin2024paintings}), and AIGC images (RichHF-18K \cite{liang2024rich}, SAC \cite{pressmancrowson2022}). We employ the Spearman Rank Correlation Coefficient (SRCC) and Pearson Linear Correlation Coefficient (PLCC) to measure model performance.

\textbf{Zero-Shot Performance:}
We first utilize DeCLIP's vision-language alignment capability in the decoupled perceptual space for zero-shot evaluation, and compare it with the vanilla CLIP. The results are summarized in \cref{tab2}. It is observed that DeCLIP significantly outperforms CLIP across all two perceptual quality evaluation tasks. Notably, for the TQA task, DeCLIP demonstrates a more pronounced improvement on authentic distortions compared to synthetic and AI-generated. In addition, DeCLIP exhibits more significant advantage when applied to natural images.

\textbf{Generalization Performance:}
The intricate entanglement of semantics and perception poses significant challenge to the generalization ability of perceptual quality evaluation models. To evaluate the generalization of DeCLIP on perceptual evaluation, we conduct cross-database experiment. We compare DeCLIP with state-of-the-art models in technical quality and aesthetic quality assessments. We utilize CoOP \cite{zhou2022learning} to train a textual prompt vector for the parameter-frozen DeCLIP, enabling cross-database testing. The experimental results are listed in \cref{tab3}. The results demonstrate that DeCLIP achieves the best performance. Notably, DeCLIP demonstrates a significant advantage in the TQA tests from real distortion (KonIQ-10k) to synthetic distortion (KADID), as well as in the AQA tests transitioning from natural images (AVA) to artistic images (APDD). This further validates the effectiveness of decoupled perceptual representations.

\begin{table}[t]
	\centering
	\small
	\begin{tabular}{cccc}
		\toprule
		Training   & \multicolumn{3}{c}{KonIQ-10k}                    \\ 
		\midrule
		Testing    & SPAQ           & AGIQA          & KADID          \\ 
		\midrule
		DBCNN \cite{8576582}      & 0.806          & 0.641          & 0.484          \\
		MetaIQA \cite{zhu2020metaiqa}    & 0.841          & 0.552          & 0.554          \\
		HyperIQA \cite{su2020blindly}   & 0.788          & 0.640          & 0.468          \\
		MUSIQ \cite{ke2021musiq}      & 0.863          & 0.630          & 0.556          \\
		CLIP-IQA \cite{wang2023exploring}  & 0.864          & 0.685          & 0.654          \\
		LIQE \cite{zhang2023blind}       & 0.833          & 0.708          & 0.662          \\ 
		\midrule
		{DeCLIP} & {\textbf{0.867}} & {\textbf{0.711}} & {\textbf{0.741}} \\ 
		\midrule \midrule
		Training   & \multicolumn{3}{c}{AVA}                          \\ 
		\midrule
		Testing    & AADB           & PARA           & APDD           \\ 
		\midrule
		NIMA \cite{talebi2018nima}       & 0.471          & 0.626          & 0.220          \\
		PA-IAA \cite{li2020personality}     & 0.527          & 0.655          & 0.428          \\
		TANet \cite{he2022rethinking}      & 0.328          & 0.471          & 0.351          \\
		TAVAR \cite{10054147}      & 0.480          & 0.652          & 0.413          \\
		VILA \cite{ke2023vila}       & 0.548          & 0.649          & 0.426          \\
		TMCR \cite{yang2024semantics}       & 0.507          & 0.622          & 0.483          \\ 
		\midrule
		{DeCLIP} & {\textbf{0.583}} & {\textbf{0.669}} & {\textbf{0.549}} \\ 
		\bottomrule
	\end{tabular}
	\caption{Comparison of cross-dataset performance on technical and aesthetic quality assessment. The SRCC index is reported. Bolded numbers indicate the best performance.}
	\label{tab3}
\end{table}

\textbf{Fine-Grained Attribute Prediction:}
In addition to the overall perception of an image, we also evaluate the zero-shot performance of DeCLIP on predicting fine-grained attributes. Specifically, we conduct experiments on two databases, SPAQ \cite{fang2020perceptual} and PARA \cite{yang2022personalized}, which provide annotations for various attributes, including quality attributes such as noise and contrast, as well as aesthetic attributes like composition and depth-of-field. The results are summarized in \cref{tab4}. It is observed from the table that DeCLIP outperforms CLIP significantly in various quality and aesthetic attributes. Particularly, DeCLIP demonstrates an approximate 20\% improvement over CLIP for the quality attributes of colorfulness, contrast, and noise. Similarly, DeCLIP exhibits a significant advantage for aesthetic attributes of depth-of-field and light. These results further demonstrate the superiority of DeCLIP on understanding fine-grained attributes, which could be highly desired in evaluation interpretability.

\subsection{Performance on Conditional Image Generation}
The decoupled perceptual and semantic vision-language alignment sub-spaces of DeCLIP provide a flexible control foundation for image generation tasks. Specifically, the decoupled visual features, in conjunction with complementary semantic or perceptual text descriptions, can serve as accurate prompts to guide the corresponding aspects of generated images. To demonstrate the capability of DP-Adapter in controlling the semantic and perceptual aspects in image generation, we conduct qualitative and quantitative comparisons with the state-of-the-art multi-modal generative models, including IP-Adapter \cite{ye2023ip}, Versatile Diffusion \cite{xu2023versatile}, ControlNet Shuffle \cite{zhang2023adding}, BLIP-Diffusion \cite{li2024blip}, and InstantStyle \cite{wang2024instantstyle}.

\begin{table}[t]
	\centering
	\setlength{\tabcolsep}{2pt}
	\scriptsize
	\begin{tabular}{@{}cccc||cccc@{}}
		\toprule
		\multicolumn{4}{c||}{\textit{TQA Attribute (SPAQ database)}}                          & \multicolumn{4}{c}{\textit{AQA Attribute (PARA database)}}                            \\ \midrule \midrule
		\multicolumn{1}{c|}{Attribute}                & Model  & SRCC   & PLCC   & \multicolumn{1}{c|}{Attribute}                   & Model  & SRCC  & PLCC  \\ \midrule
		\multicolumn{1}{c|}{\multirow{2}{*}{Brightness}}   & CLIP   & 0.439  & 0.437  & \multicolumn{1}{c|}{\multirow{2}{*}{Composition}}     & CLIP   & 0.274 & 0.326 \\
		\multicolumn{1}{c|}{}                              & {DeCLIP} & \textbf{0.584}  & \textbf{0.529}  & \multicolumn{1}{c|}{}                                 & {DeCLIP} & \textbf{0.369} & \textbf{0.431} \\ \midrule
		\multicolumn{1}{c|}{\multirow{2}{*}{Colorfulness}} & CLIP   & 0.278  & 0.251  & \multicolumn{1}{c|}{\multirow{2}{*}{Color}}           & CLIP   & 0.444 & 0.471 \\
		\multicolumn{1}{c|}{}                              & {DeCLIP} & \textbf{0.576}  & \textbf{0.412}  & \multicolumn{1}{c|}{}                                 & {DeCLIP} & \textbf{0.530} & \textbf{0.563} \\ \midrule
		\multicolumn{1}{c|}{\multirow{2}{*}{Contrast}}     & CLIP   & -0.135 & -0.154 & \multicolumn{1}{c|}{\multirow{2}{*}{Depth-of-Field}}  & CLIP   & 0.113 & 0.172 \\
		\multicolumn{1}{c|}{}                              & {DeCLIP} & \textbf{0.311}  & \textbf{0.239}  & \multicolumn{1}{c|}{}                                 & {DeCLIP} & \textbf{0.374} & \textbf{0.445} \\ \midrule
		\multicolumn{1}{c|}{\multirow{2}{*}{Noise}}    & CLIP   & 0.195  & 0.112  & \multicolumn{1}{c|}{\multirow{2}{*}{Light}}           & CLIP   & 0.488 & 0.536 \\
		\multicolumn{1}{c|}{}                              & {DeCLIP} & \textbf{0.573}  & \textbf{0.286}  & \multicolumn{1}{c|}{}                                 & {DeCLIP} & \textbf{0.591} & \textbf{0.622} \\ \midrule
		\multicolumn{1}{c|}{\multirow{2}{*}{Sharpness}}    & CLIP   & 0.638  & \textbf{0.395}  & \multicolumn{1}{c|}{\multirow{2}{*}{Object emphasis}} & CLIP   & 0.259 & 0.281 \\
		\multicolumn{1}{c|}{}                              & {DeCLIP} & \textbf{0.794}  & {0.256}  & \multicolumn{1}{c|}{}                                 & {DeCLIP} & \textbf{0.325} & \textbf{0.317} \\ \bottomrule
	\end{tabular}
	\caption{Performance comparison on attribute prediction (zero-shot). Bolded numbers indicate better performance.}
	\label{tab4}
\end{table}

\begin{figure*}[t]
	\centering
	\includegraphics[width=0.93\linewidth]{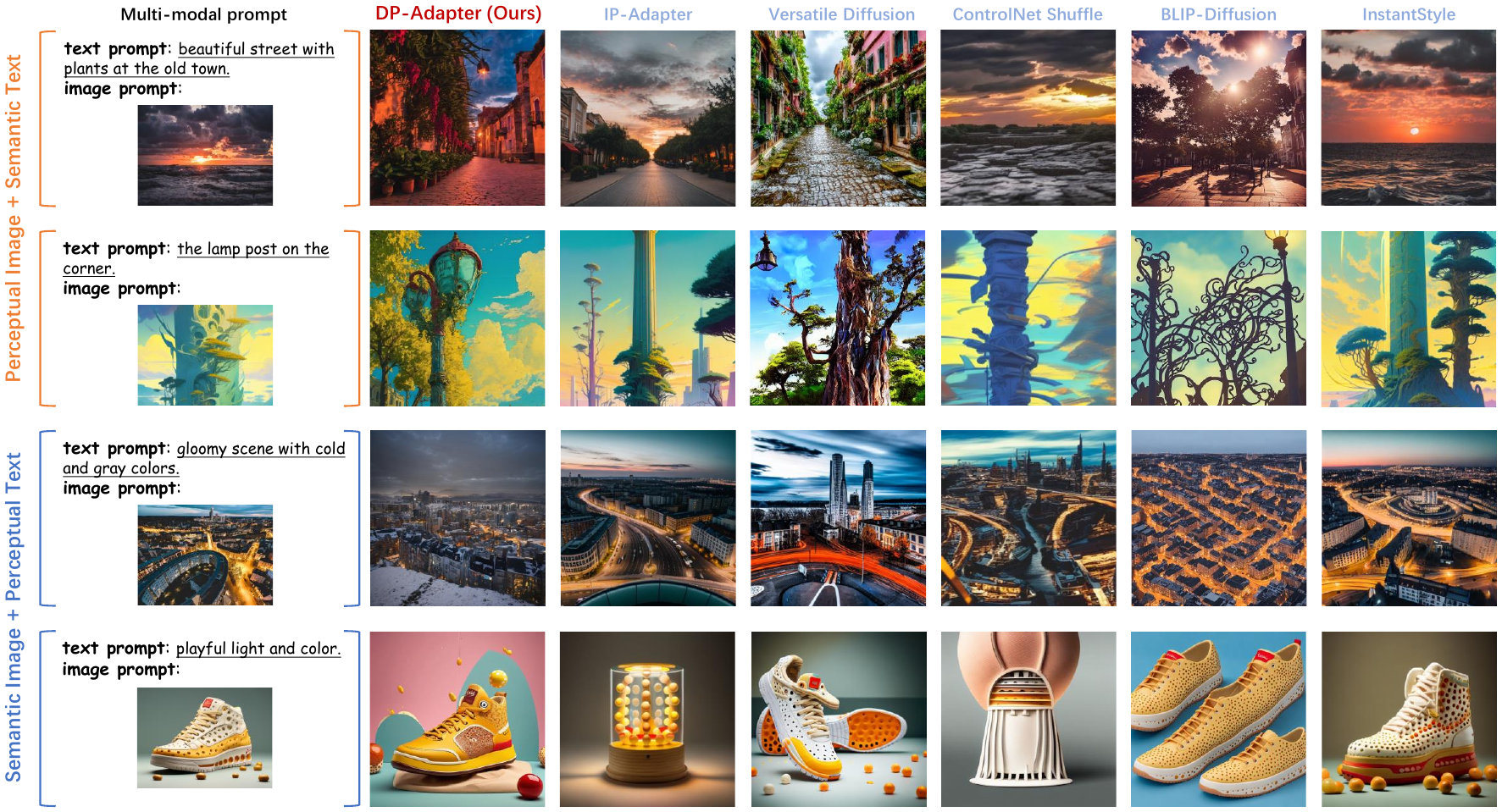}
	
	\caption{The visual comparison of the DP-Adapter with other methods conditioned on different perceptual and semantic prompts.}
	\label{fig4}
\end{figure*}

\textbf{Qualitative Results:}
\cref{fig4} shows a qualitative comparison between our proposed DP-Adapter and five popular multi-modal generative models. We compare the generative performance of all models under two different control types, using identical image and text prompts.

 $Perceptual$ $Image$ $+$ $Semantic$ $Text$ signifies that the image prompt provides perceptual reference, while the text prompt offers semantic reference for image generation. As illustrated in the top two rows in \cref{fig4}, DP-Adapter demonstrates superior generative performance compared to the other models. The compared models exhibit varying degrees of `mismatch' defects. Notably, ControlNet Shuffle and InstantStyle do not incorporate semantic elements from the text prompt such as `street' and `lamp'. Versatile Diffusion exhibits a significant perceptual discrepancy with the image prompt. The  substantial semantic differences between image and text prompts severely impact generative outcomes. In contrast, DP-Adapter effectively mitigates this issue by utilizing decoupled representations. 

$Semantic$ $Image$ $+$ $Perceptual$ $Text$ indicates that the image prompt provides semantic reference, while the text prompt provides perceptual reference for generating images. Similarly, from the examples in the bottom two rows of \cref{fig4}, we observe that DP-Adapter also yields the best generation results. An interesting finding is that the word `light' has different interpretations from semantic and perceptual perspectives. Both IP-Adapter and ControlNet Shuffle interpret it as a semantic element, which significantly deviates from the intended control objectives. These results demonstrate the advantage of DP-Adapter in controlling  both perceptual and semantic aspects in image generation.

\textbf{Quantitative Results:}
We further conduct a user study to evaluate the semantic and perceptual control capabilities of DP-Adapter. Specifically, we collected 100 pairs of image-text as conditions for image generation. The images are sourced from existing datasets \cite{LITE}, including natural, artistic, and AIGC. The textual descriptions are categorized into two types: perception and semantics. Some of the descriptions originate from established datasets (e.g., COCO Captions \cite{chen2015microsoft}), while others are generated by Large Language Model. These text prompts are then randomly combined with the selected image prompts to produce 100 image-text pairs, which are then used for image generation with DP-Adapter and the compared models.

\begin{table}[t]
	\centering
	\small
	\begin{tabular}{c|cc}
		\toprule
		Metrics             & CC             & OF             \\ \midrule
		IP-Adapter \cite{ye2023ip}          & 0.628          & 0.536          \\
		Versatile Diffusion \cite{xu2023versatile} & 0.521          & 0.400          \\
		ControlNet Shuffle \cite{zhang2023adding}  & 0.465          & 0.293          \\
		BLIP-Diffusion \cite{li2024blip}      & 0.436          & 0.292          \\
		InstantStyle \cite{wang2024instantstyle}        & 0.464          & 0.288          \\  \midrule
		DP-Adapter          & \textbf{0.696} & \textbf{0.641} \\ \bottomrule
	\end{tabular}
	\caption{Quantitative comparisons from human on generative performance. Best result is marked in bold.}
	\label{tab5}
\end{table}

We recruited ten volunteers to perform the user study. Specifically, they are asked to score (1-5) the generated images from three dimensions: Semantic Consistency (SC), Perceptual Consistency (PC), and Overall Feeling (OF). The first two are used to assess the alignment between the generated images and the corresponding semantic (e.g., objects) and perceptual (e.g., color) conditions. For OF, factors such as adherence to conditions, presence of defects, and aesthetic appeal are considered. Finally, we calculate the average score for each dimension and derive an overall Condition Consistency (CC) by averaging SC and PC scores, to assess the generative performance. The results are summarized in \cref{tab5}. The CC and OF metrics indicate that the proposed DP-Adapter achieves the highest generation quality. This is attributed to the decoupled vision-language representations.

%% file: sec/5_conclusion.tex
\section{Conclusion}
In this paper, we propose a new multi-modal framework for disentangled representation learning. Our method leverages MLLMs-assisted text disentanglement to learn perception-specific and semantics-specific visual representations, thereby achieving fine-grained vision-language alignment. Comprehensive experiments conducted on both image quality assessment and conditional image generation tasks demonstrate the remarkable zero-shot generalization capability of these representations. While very encouraging results have been achieved in this work, the decoupled features still hold significant potential for further enhancement and application, especially in perception-related tasks like image enhancement and restoration.